\documentclass{article}
\usepackage{spconf,amsmath,graphicx}
\usepackage{amsthm,amssymb}
\usepackage{cleveref}
\usepackage{url}
\usepackage{graphicx}
\usepackage{enumitem}
\usepackage{makecell}
\usepackage{multirow}
\usepackage{array}
\usepackage{booktabs}

\title{Improving Sentence Similarity Estimation for \\ Unsupervised Extractive Summarization}
\name{Shichao Sun$^{1}$ \qquad Ruifeng Yuan$^{1}$ \qquad Wenjie Li$^{1}$ \qquad Sujian Li$^{2}$}
\address{$^{1}$~The Hong Kong Polytechnic University \qquad  $^{2}$~Peking University \\
\{csssun, csryuan, cswjli\}@comp.polyu.edu.hk \qquad lisujian@pku.edu.cn
}
\begin{document}
%
\maketitle
\begin{abstract}
Unsupervised extractive summarization aims to extract salient sentences from a document as the summary without labeled data. Recent literatures mostly research how to leverage sentence similarity to rank sentences in the order of salience. However, sentence similarity estimation using pre-trained language models mostly takes little account of document-level information and has a weak correlation with sentence salience ranking. In this paper, we proposed two novel strategies to improve sentence similarity estimation for unsupervised extractive summarization. We use contrastive learning to optimize a document-level objective that sentences from the same document are more similar than those from different documents. Moreover, we use mutual learning to enhance the relationship between sentence similarity estimation and sentence salience ranking, where an extra signal amplifier is used to refine the pivotal information. Experimental results demonstrate the effectiveness of our strategies.~\footnote{Our code is available at:~\url{https://github.com/ShichaoSun/SS4Sum}}

\end{abstract}
\section{Introduction}
Text summarization aims to condense a long document into its shorter version while preserving salient content. Existing methods can be divided into two paradigms, i.e., abstractive and extractive. Abstractive methods generate a summary word by word. Extractive methods select salient sentences from a document as the summary. Modern neural network based approaches~\cite{ see-etal-2017-get, zhong-etal-2020-extractive, liu-etal-2022-brio} have achieved promising results, which heavily rely on the large-scale annotated corpus. However, it is unrealistic to expect large-scale and high-quality annotated corpus to be available all the time. It therefore comes as no surprise that unsupervised summarization has attracted much attention~\cite{mihalcea-tarau-2004-textrank, zheng-lapata-2019-sentence, xu-etal-2020-unsupervised, yang-etal-2020-ted, liang-etal-2021-improving, liu2021unsupervised}. Most attempts are extractive since it is obviously difficult to generate the summary sentences without any reference summary.  

Unsupervised extractive summarization is commonly graph-based~\cite{zheng-lapata-2019-sentence, liang-etal-2021-improving, liu2021unsupervised}. These methods contain two stages. The first stage is to obtain a sentence encoder, which encodes a sentence into an embedding. Pre-trained language models like BERT~\cite{devlin-etal-2019-bert} are used in this stage. The second stage is to calculate the salience score via the sentence embedding, and sentences with the highest scores are selected as a summary. In this stage, a document is represented as a graph, where nodes represent sentences. The weight of an edge is the similarity of two adjacent nodes (sentences), which can be estimated by dot product or cosine distance using the sentence embedding. Then the node centrality is calculated as the salience score of a sentence. Most existing methods put more effort into this step while they commonly used pre-trained language sentence encoders. PacSum~\cite{zheng-lapata-2019-sentence} added the direction information to the degree centrality. FAR~\cite{liang-etal-2021-improving} incorporated the facet information into PacSum, and DASG~\cite{liu2021unsupervised} augmented PacSum with the distance information. 

However, there is a gap between training a sentence encoder and estimating the similarity of two sentences using dot product or cosine distance. The training objectives of a sentence encoder are commonly masked language model~\cite{liu2021unsupervised} and neighboring sentence prediction~\cite{zheng-lapata-2019-sentence, liang-etal-2021-improving, liu2021unsupervised}. There is not any explicit correlation between sentence similarity estimation and dot product or cosine distance of two sentence embeddings. These pre-trained models are not expected to well estimate sentence similarity using dot product or cosine distance. It is also intuitive that by only considering a sentence and its neighboring sentences, the sentence embedding can hardly capture the document-level similarity that sentences from the same document are more similar than those from different documents. Besides, TF-IDF can even outperform BERT in unsupervised extractive summarization as shown in~\Cref{tab:result}. It indicates that there is a weak relationship between sentence similarity estimation and sentence salience ranking. This motivated us to explore how to improve sentence similarity estimation for unsupervised extractive summarization.

To address the above issues, we propose two novel strategies to train the sentence encoder. To enable the pre-trained models to get aware of document-level information, we use contrastive learning to optimize a document-level objective that sentences from the same document are more similar than those from different documents. To build the relationship between similarity estimation and dot product of two sentence embeddings, we define the above sentence similarity as the dot product of two sentence embeddings. Moreover, we use mutual learning~\cite{zhang2018deep} to enhance the relationship between sentence similarity estimation and sentence salience ranking. An extra amplifier called Deep Differential Amplifier~\cite{jia-etal-2021-deep} is used to refine the pivotal information. It learns from the coarse-grained pivotal information that the top 40\% ranked sentences are marked as salient sentences, and the bottom 40\% ranked sentences are marked as unimportant sentences, where the salience scores are estimated using our sentence encoder. It will output the fine-grained pivotal information that top 3 sentences are marked as salient sentences, and other sentences are marked as unimportant sentences. This fine-grained pivotal information is used to adjust our sentence salience ranking. It means that under the mutual learning framework, the Deep Differential Amplifier learns the salience scores calculated by our sentence similarity estimation. Meanwhile our calculated salience scores are supervised by the predicted results of the Deep Differential Amplifier. 

We conduct experiments on two datasets, i.e.,~\textbf{NYT}~\cite{sandhaus2008new} and~\textbf{CNNDM}~\cite{hermann2015teaching}. Experimental results show that our sentence similarity estimation beats other similarity estimation methods in unsupervised extractive summarization. The ablation study  demonstrates the effectiveness of our strategies.

\section{Method}

\subsection{Similarity Estimation}
\label{sec:sim}
Let $D$ denote the document consisting of a sequence of sentences $\{s_1,s_2,\cdots,s_n\}$. And $e_{ij}$ denotes the similarity for each sentence pair ($s_i$, $s_j$). Their similarity is dot product of their sentence embeddings, which is calculated as follows:
\begin{equation}
\label{eq:eij}
e_{ij} = v_i^{\top} v_j 
\end{equation}
where $v_i$ is the embedding of sentence $s_i$ and $v_j$ is the embedding of sentence $s_j$. Note that the sentence embedding is generated by a sentence encoder like BERT. 

\subsection{Contrastive Learning}
\label{sec:con_learning}
Contrastive learning is used to optimize the novel objective that the similarity of sentences from the same document are higher than the similarity of sentences from different documents, because sentences of a document contribute to one core. Specifically, we utilize the contrastive learning to increase the dot product of sentences from the same document. Meanwhile, we decrease the dot product of sentences from different sentences (within the same batch). This is because in unsupervised extractive summarization dot product is used to estimate sentence similarity. For the sentences $\{s_1,s_2,\cdots,s_n\}$ in the same batch, where we make $s_{2i}$ and $s_{2i+1}$ belong to the same document, and $s_{2i}$ and $s_{j}$ ($j \neq 2i$ and $j \neq 2i + 1 $) belong to different documents. The contrastive learning loss  $L_{con}$ can be calculated as follows:
\begin{equation}
    L_{con} = -\log\frac{\exp(v_{2i}^{\top} v_{2i+1} / \tau)}{\sum\limits_{j \neq 2i} \exp(v_{2i}^{\top} v_{j} / \tau) }
\end{equation}
where $\tau$ is a scalar temperature parameter. Note that this loss function can directly optimize sentence similarity. 

\subsection{Salience Score}
\label{sec:salience_score}
The salience score is used to rank sentences of a document, which is calculated by a graph-based algorithm. A document is represented as a graph in which nodes correspond to sentences, and each edge between two sentences is weighted by their similarity. Degree Centrality (DC) is a popular method for calculating the salience score. It is based on the hypothesis that a salient sentence should be similar to the rest sentences in the document. The DC of sentence $s_i$ from a document containing $n$ sentences is calculated as follows:
\begin{equation}
\label{eq:dc}
    {\rm DC}(s_i) = \sum_{j \in \{0, \cdots, i-1, i+1, \cdots, n \}} e_{ij}
\end{equation}

\subsection{Mutual Learning}
\label{sec:mul_learning}

\begin{figure*}[htb] 
\centering 
\includegraphics[width=0.85\textwidth]{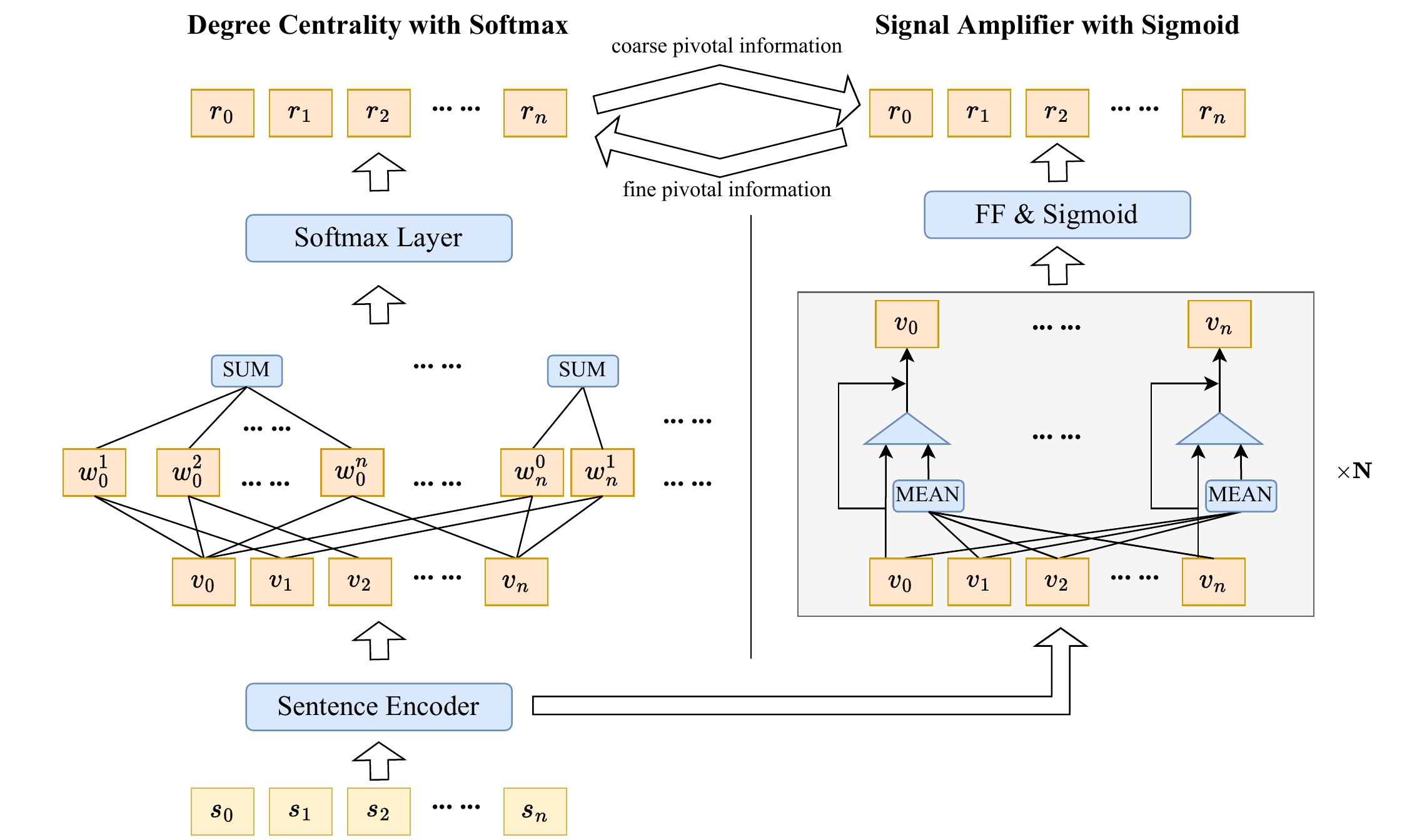} 
\caption{Mutual Learning for Sentence Embedding Learning.} 
\label{fg:mutual_learning}
\end{figure*}
Mutual learning is used to enhance the relationship between sentence similarity estimation and sentence salience ranking by an extra signal amplifier as illustrated in~\Cref{fg:mutual_learning}. The left part of this figure is our sentence salience ranking (degree centrality) as described in~\Cref{sec:salience_score} with an extra Softmax layer. The output is the ranking score $r_i$ of each sentence $s_i$. The right part of this figure is a signal amplifier called Deep Differential Amplifier, which is able to amplify the salient signal between the current sentence and other sentences. A feedforward neural network with a Sigmoid layer is used to convert the salient signal to ranking score $r_i$ of each sentence $s_i$. Note that this signal amplifier is used to refine the pivotal information. This signal amplifier will be supervised by the coarse-grained pivotal information that the top 40\% ranked sentences are marked as salient sentences, and the bottom 40\% ranked sentences are marked as unimportant sentences, where the salience scores are estimated using degree centrality as shown in the left part of this figure. And this signal amplifier will output the fine-grained score for each sentence that the top 3 sentences are marked as salient sentences, and other sentences are marked as unimportant sentences. This is motivated by the strong ability of generalization. Conversely, this fine-grained score will be used to adjust the salience ranking score of the model in the left part of this figure. We jointly train degree centrality with a softmax layer and the signal amplifier with a sigmoid layer together to construct a mutual learning variant. During this process, the sentence similarity estimation is indirectly optimized to adapt to sentence salience ranking.

The left part of~\Cref{fg:mutual_learning} is our sentence salience ranking (degree centrality) as described in~\Cref{sec:salience_score} with an extra Softmax layer, which is used to get normalized salience score. It can be calculated as follows:
\begin{equation}
\label{eq:r_i}
    r_i = \frac{\exp(\frac{{\rm DC}(s_i)}{\tau (n - 1)})}{\sum\limits_{j \in D} \exp(\frac{{\rm DC}(s_j)}{\tau (n - 1)})}
\end{equation}
where $n$ is the number of sentences in a document $D$ and $\tau$ is a scalar temperature parameter. This salient score will be optimized by the loss function as follows:
\begin{equation}
    L_{dc} = - \log \sum\limits_{i \in C} \frac{r_i}{\sum\limits_{j \in D} r_j}
\end{equation}
where $C$ consists of the salient sentences that the signal amplifier (right part) selects as a summary, i.e., the top 3 salient sentences. It means to train the sentence encoder by using the fine-grained pivotal information from the amplifier. 

The right part is the state-of-the-art supervised extractive summarization method, Deep Differential Amplifier~\cite{jia-etal-2021-deep}. The gray area describes the process of amplifying salient signal, which is iterated for $N$ times as follows:
\begin{equation}
\begin{split}
        F(v_i) &= {\rm MLP}(v_i - {\rm mean}(\{v_j \mid j \neq i\})) \\
        v_i &= F(v_i) + v_i
\end{split}
\end{equation}
where $v_i$ is the sentence embedding generated by the sentence encoder with the input of a sentence $s_i$, ${\rm MLP}$ is a multilayer perceptron with two layers and the ReLU activation function. 
And the salience score $r_i$ is calculated as follows:
\begin{equation}
    r_i = {\rm sigmoid}(\mathbf{w^{\top}} v_i)
\end{equation}
where $\mathbf{w}$ is the trainable parameters. It is optimized by using the binary cross entropy as follows:
\begin{equation}
    L_{amp} = - y_i \log (r_i) - (1 - y_i) \log (1 - r_i)  
\end{equation}
where $y_i$ is 1 if the sentence $s_i$ is one of the top-k ranked sentences, and $y_i$ is 0 if sentence $s_i$ is one of the bottom-k ranked sentences according to the salience scores. The salience scores are calculated according to \Cref{eq:r_i}. $k$ is around 40\% sentence number of a document. This means to train the amplifier by using the coarse-grained pivotal information.

\subsection{Training Objective}
\label{sec:objective}
Finally, the sentence encoder will be trained by summing the above three loss $L$ as follows:
\begin{equation}
    L = L_{con} + L_{dc} + L_{amp}
\end{equation}

\begin{table*}[tbp]
\centering
\caption{The Rouge scores on CNNDM and NYT. }
\begin{tabular}{lcccccc}
\toprule
  & \multicolumn{3}{c}{\textbf{CNNDM}} & \multicolumn{3}{c}{\textbf{NYT}} \\
& \multicolumn{1}{c}{Rouge-1} & \multicolumn{1}{c}{Rouge-2} & \multicolumn{1}{c}{Rouge-L} & \multicolumn{1}{c}{Rouge-1} & \multicolumn{1}{c}{Rouge-2} & \multicolumn{1}{c}{Rouge-L} \\
\hline
TF-IDF  & 33.00  & 11.70 & 29.50 & 33.20  & 13.10 & 29.00 \\
BERT  & 28.33 & 8.29  & 25.38 & 25.87 & 7.29  & 22.28 \\
SimCSE  & 31.98  & 10.42 & 28.52 & 30.55  & 10.26 & 26.38 \\
PacSum BERT  & 31.42  & 10.07 & 28.00 & 30.99  & 10.28 & 26.81 \\
SimBERT (ours)   & \textbf{35.41}  & \textbf{13.18}   & \textbf{31.75} & \textbf{35.20}  & \textbf{14.75}   & \textbf{30.97} \\
\bottomrule
\end{tabular}
\label{tab:result}
\end{table*}

\section{Experiment and Analysis}
\subsection{Datasets}
We evaluate our sentence similarity estimation by testing whether it can improve the performance for unsupervised extractive summarization on two summarization datasets, i.e.,~\textbf{NYT}~\cite{sandhaus2008new} and~\textbf{CNNDM}~\cite{hermann2015teaching}. Our sentence encoder is trained by using the input documents of CNNDM training dataset and NYT training dataset. The final training dataset contains 322,828 documents with more than five sentences. For the sake of quickly validating, we randomly select 500 samples from CNNDM validation dataset and 500 samples from NYT validation dataset. 

\subsection{Baselines}
We choose four different sentence representation methods as baselines. The first one is TF-IDF, whose value of the corresponding dimension is the tf (term frequency) times the idf (inverse document frequency) of the word. The second one uses the original BERT to encode the sentence. The third one is SimCSE~\cite{gao-etal-2021-simcse}, which takes an input sentence and predicts itself in a contrastive learning objective, with only dropout used as noise. The fourth one (PacSum BERT) comes from the~\cite{zheng-lapata-2019-sentence}, which captures semantic information by distinguishing context sentences from other sentences. These sentence representations are used to calculate the degree centrality (\Cref{eq:dc}), and the top 3 sentences are selected as a summary.

\subsection{Automatic Evaluation}

We automatically evaluate summary quality using Rouge~\cite{lin-2004-rouge}, and the experimental results on CNNDM and NYT are presented in~\Cref{tab:result}. It shows that our sentence embedding (SimBERT) achieved the best performance on CNNDM and NYT, so it can be proved that our sentence similarity estimation can be more suitable for unsupervised extractive summarization. It can be attributed to the novel training objectives of explicitly capturing document-level sentence similarity and enhancing the relationship between sentence similarity and sentence salience ranking.

Besides, it should be noted that TF-IDF can perform better than the other BERT variants except SimBERT. It is counterintuitive because the pre-training based representation (like BERT) always outperforms statistic based representation (like TF-IDF). It indicates that current pre-training based sentence encoder is not good at capturing sentence similarity. This result motivated us to explore how to estimate sentence similarity for unsupervised extractive summarization.

\begin{table}[tbp]
\centering
\caption{The results of ablation study.}
\begin{tabular}{lccc}
\toprule
 & \multicolumn{3}{c}{\textbf{CNNDM and NYT}} \\
 & \multicolumn{1}{c}{R1} & \multicolumn{1}{c}{R2} & \multicolumn{1}{c}{RL} \\
\hline
SimBERT  & 35.36  & 13.62 & 31.54 \\
\quad - mutual learning  & 31.68 & 10.50  & 28.04 \\
\quad - contrastive learning  & 27.54  & 7.96 & 24.42 \\
\bottomrule
\end{tabular}
\label{tab:ablation}
\end{table}

\subsection{Ablation Study}

The ablation study is conducted on the merged testing dataset of CNNDM and NYT to evaluate the contribution of mutual learning ($L_{dc}$ and $L_{amp}$) and contrastive learning ($L_{con}$). The Rouge scores are given in~\Cref{tab:ablation}. It tells that mutual learning and contrastive learning are complementary since we can achieve better results by using both of them. Besides, it can be found that the performance will degrade a lot without contrastive learning. It can indicate that the sentence similarity in the document level is important. The degradation without mutual learning can show that there is a weak relationship between sentence similarity and sentence salience ranking.

\section{Conclusion}
In this paper, we proposed two novel strategies to improve sentence similarity estimation for unsupervised extractive summarization. We use contrastive learning to optimize a document-level objective that sentences from the same document are more similar than those from different documents. Moreover, we use mutual learning to enhance the relationship between sentence similarity estimation and sentence salience ranking, where an extra signal amplifier is used to refine the pivotal information. Experimental results show that our sentence similarity estimation beats other similarity estimation methods in unsupervised extractive summarization. The ablation study demonstrates the effectiveness of our strategies. 

\section{Acknowledgement}
This paper was supported by Research Grants Council of Hong Kong (PolyU/15203617 and PolyU/5210919), National Natural Science Foundation of China (61672445, 62076212).

\bibliographystyle{IEEEbib}
\bibliography{refs}

\end{document}